\documentclass[runningheads]{llncs}
\usepackage{graphicx}
\usepackage{xcolor}
\usepackage{comment}
\usepackage{bm}
\usepackage{amssymb}
\usepackage{amsmath}

\begin{document}
\title{Synth-by-Reg (SbR): Contrastive learning for synthesis-based registration of paired images} 
\titlerunning{Synth-by-Reg (SbR): Contrastive learning for synthesis-based registration}

\author{
Adri\`a Casamitjana\inst{1*}  \and
Matteo Mancini \inst{2,3,4} \and
Juan Eugenio Iglesias\inst{1,5,6}}

\authorrunning{A. Casamitjana et al.}

\institute{
Center for Medical Image Computing, University College London, UK \\
\email{a.casamitjana@ucl.ac.uk} 
\and
Department of Neuroscience, University of 
Sussex, Brighton, UK \and
NeuroPoly Lab, Polytechnique Montreal, Canada.  \and
CUBRIC, Cardiff University, UK \and
Martinos  Center  for  Biomedical  Imaging,  MGH  and  Harvard  Medical School, USA \and
Computer  Science  and  AI  Laboratory,  Massachusetts  Institute  of  Technology, USA
}

\maketitle
\begin{abstract}
Nonlinear inter-modality registration is often challenging  due to the lack of objective functions that are good proxies for alignment. Here we propose a synthesis-by-registration method to convert this problem into an easier intra-modality  task. We introduce a registration loss for weakly supervised image translation between domains that does not require perfectly aligned training data. This loss capitalises on a registration U-Net with frozen weights, to drive a synthesis CNN towards the desired  translation. We complement this loss with a structure preserving constraint based on contrastive learning, which prevents blurring and content shifts due to overfitting. We apply this method to the registration of histological sections to MRI slices, a key step in 3D histology reconstruction. Results on two public datasets show improvements over registration based on mutual information (13\% reduction in landmark error) and synthesis-based algorithms such as CycleGAN (11\% reduction), and are comparable to registration with label supervision. Code and data are publicly available at \url{https://github.com/acasamitjana/SynthByReg}.
\keywords{Image synthesis  \and Inter-modality registration \and Deformable registration \and Contrastive estimation}
\end{abstract}

\section{Introduction}

Image registration is a crucial step  to spatially relate information from different medical images. Unpaired  registration aligns images of different subjects into a common space to perform subsequent analysis (e.g., population studies~\cite{fonov2011unbiased}, voxel-based morphometry~\cite{ashburner2000voxel}, or multi-atlas segmentation~\cite{rohlfing2004evaluation,iglesias2015multi}). On the other hand, paired  registration  aligns different images from the same anatomy and finds application in image guided intervention (e.g., MR-CT in the prostate ~\cite{hu2012mr}); patient follow-up (e.g., pre- and post-operative scans~\cite{kwon2013portr}); or longitudinal~\cite{reuter2012within} and multimodal studies (e.g., 3D histology reconstruction with MRI~\cite{pichat2018survey}).

Registration is often cast as an optimisation problem where a source image is deformed towards a target image such that it maximises a similarity metric of choice. Classical registration methods solve this problem independently for every pair of images with standard iterative optimisers~\cite{sotiras2013deformable}. Modern learning approaches predict a deformation directly from a pair of images using a convolutional neural network (CNN). Supervised learning methods use ground truth deformation fields in training, either synthetic~\cite{sokooti2017nonrigid} or derived from manual segmentations~\cite{cao2017deformable}. These have been superseded by unsupervised methods, in which CNNs are trained to optimise metrics like those used in classical registration, e.g., sum of squared differences (SSD) or  local normalised cross-correlation (LNCC)~\cite{balakrishnan2019voxelmorph,de2017end}, without wasting  capacity in regions without salient features.

Widespread similarity functions like SSD or LNCC are well suited for intra-modality registration problems.
However, the difficulty of designing accurate similarity functions across modalities hampers inter-modality registration. Mutual information (MI) is often used~\cite{maes2003medical} but with unsatisfactory results in the nonlinear case, due to the excessive flexibility of the model~\cite{iglesias2013synthesizing}. 
Other metrics used in inter-modality registration are the Modality-Independent Neighbourhood Descriptor (MIND,~\cite{heinrich2012mind}, based on local patch similarities) or adversarial losses measuring  whether two images are well aligned or not~\cite{fan2019adversarial}. MIND is sensitive to initial alignment, bias field or rotations depending on the neighbourhood size, while adversarial losses are prone to missing local correspondences.

An alternative to inter-modality registrationis to convert the problem into an intra-modality task using a registration-by-synthesis framework: image-to-image (I2I) translation  is first  used to synthesise new source images with the target contrast, and then intra-modality  registration (which is more accurate) is performed in the target domain. With accurate image synthesis, the errors introduced by the translation are outweighed by the improvement in registration~\cite{iglesias2013synthesizing}. In unsupervised synthesis, cycle-consistent generative adversarial networks (CycleGAN) can be used~\cite{tanner2018generative,wei2019synthesis}, but they lack structural consistency across views and may generate artefacts  due to overfitting (e.g., flip contrast or even deform images). To mitigate this issue,  additional losses between the original and synthetic images have been proposed, e.g., segmentation losses~\cite{huo2018adversarial} or inter-modality similarities between the original and synthetic scans (e.g., MIND~\cite{xu2020adversarial} or  MI~\cite{wang2021dicyc}).  

Beyond CycleGAN, other approaches have attempted to enforce geometry consistency between the original and synthetic images via specific architectures or training schemes.
An I2I translation model that explicitly learns to disentangle domain-invariant (i.e., content) from domain specific features (i.e., appearance)  was proposed in~\cite{qin2019unsupervised}; the latent content features can then be used to train a registration network.
More recently, a novel training scheme that forces the translation and registration steps to be commutative (thus discouraging deformation at synthesis) has been presented~\cite{arar2020unsupervised}.
Nonetheless, GAN-based approaches are challenging to train, with well-known problems (e.g., vanishing gradients, instability~\cite{arjovsky2017towards}) and an increasing number of losses and hyperparameters. 
In this work, we turn the registration-by-synthesis framework around into a synthesis-by-registration (SbR) approach, where a registration network trained on the target domain (and frozen weights) is used in the loss for training an I2I network. This allows us to greatly simplify the  objective function and avoid potentially unstable adversarial training. Moreover, we use contrastive learning at the patch level to ensure geometric consistency. The SbR model outputs both the translated image and the deformation field. The contribution of this work is threefold: \emph{(i)}~we develop a novel registration loss for paired I2I translation; \emph{(ii)}~we adapt the contrastive PatchNCE loss~\cite{park2020contrastive} for image registration as a geometry-preserving constraint; and \emph{(iii)}~we combine \emph{(i)} and \emph{(ii)}  into an unsupervised SbR framework for inter-modality registration that does not require multiple encoders / decoders and therefore has low GPU memory requirements.


\begin{figure}[!t]
    \centering
    \includegraphics[width=0.8\linewidth]{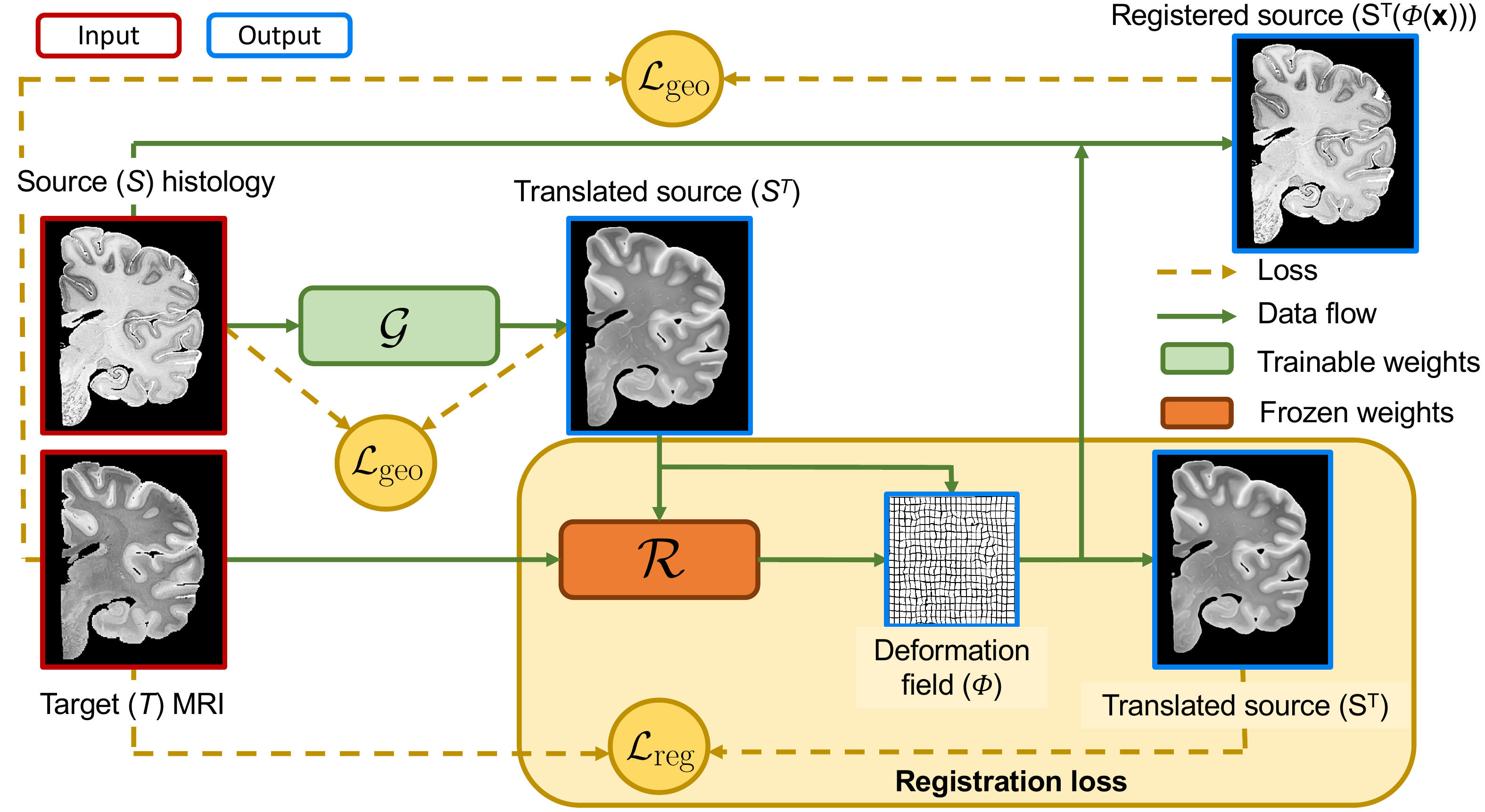}
    \caption{Overview of proposed pipeline, using histology and MRI as source and target contrasts, respectively.}
    \label{fig:Pipeline_overview}
\end{figure}

\section{Methods}

\subsection{Overview}

Let us consider two misaligned 2D images of the same anatomy (e.g., a histological section and a corresponding MRI plane): the source $S(\bm{x})$ and target $T(\bm{x})$; $\bm{x}$ represents spatial location. We further assume the availability of an intra-modality registration CNN $\mathcal{R}$ with weights $\bm{\theta}_{\mathcal{R}}$, which predicts a deformation  field $\Phi$ from two images of the target modality:
$\Phi = \mathcal{R}(T, T'; \bm{\theta}_{\mathcal{R}})$, such that $T(\bm{x})\approx T'(\Phi(\bm{x}))$. We also define an I2I translation CNN $\mathcal{G}$ (with weights $\bm{\theta}_{\mathcal{G}}$) from contrast $S$ to $T$ that regresses the image intensities: 
$S^T = \mathcal{G}(S; \bm{\theta}_{\mathcal{G}})$, such that $S^T$ resembles the anatomy in $S$, had it been acquired with modality $T$. 
The crucial observation is that, if $S^T$ is well synthesised, and $\Phi=\mathcal{R}(T, S^T; \bm{\theta}_{\mathcal{R}})$, then $T(\bm{x})\approx S^T (\Phi(\bm{x}))$. Specifically, we propose the following loss (Figure~\ref{fig:Pipeline_overview}):
\begin{equation}
    \mathcal{L}(\bm{\theta}_{\mathcal{G}}) = \mathcal{L}_{\text{reg}}(S, T; \bm{\theta}_{\mathcal{G}},\bm{\theta}_{\mathcal{R}})  + \lambda_{\text{geo}} \mathcal{L}_{\text{geo}}(S, T; \bm{\theta}_{\mathcal{G}},\bm{\theta}_{\mathcal{R}}),
    \label{eq:overview_loss}
\end{equation}
where $\mathcal{L}_{\text{reg}}$ is a ``registration loss'' measuring the similarity of $T$ and (the deformed) $S^T$,  $\mathcal{L}_{\text{geo}}$ is a geometric consistency loss that ensures that the contents of $S$ and $S^T$ are aligned, and $\lambda_{\text{geo}}$ is a relative weight. A key implicit assumption of this framework is that, because the images are paired, there exists a spatial transform $\Phi$ that aligns $S$ and $T$ well, such that the synthesis does not need to shift or blur boundaries to minimise the error; the geometric consistency loss further discourages such mistakes. Crucially, the loss in Eq.~\ref{eq:overview_loss} does not depend on $\bm{\theta}_{\mathcal{R}}$: the registration CNN is trained on the target domain and its weights are frozen, such that gradients will backpropagate through these layers to improve the synthesis. This asymmetric scheme enables us to avoid using a distribution matching loss (e.g., CycleGAN) that may produce hallucination artefacts~\cite{cohen2018distribution}.

\subsection{Intra-modality registration network}
\label{sec:intramodality}
One of the key points in SbR is the differentiable registration method used to train the image synthesis model. We use a U-Net~\cite{cciccek20163d} model (as in~\cite{dalca2018unsupervised}) that learns a diffeomorphic mapping between images from the same modality. 
The model is trained on pairs of images from the target domain and outputs a stationary velocity field (SVF), $\bm{\psi}$, at half the input resolution. Then, a scaling and squaring approach is used to integrate $\bm{\psi}$ into a half-resolution deformation field, which is linearly upsampled to obtain the final deformation $\Phi(\bm{x})$.  Training  uses LNCC  as image similarity term,  and the norm of the gradient of the SVF as regulariser:
\begin{equation}
    \mathcal{L}_{\mathcal{R}}(T,T'; \bm{\theta}_{\mathcal{R}}) = \frac{1}{|\Omega|} \sum_{\bm{x}\in\Omega} LNCC[T(\bm{x}), T'(\Phi(\bm{x}; \bm{\theta}_{\mathcal{R}}))] + 2  \lambda_{\mathcal{R}}  \| \nabla \bm{\psi}(\bm{x}; \bm{\theta}_{\mathcal{R}})  \|_2,
\end{equation}
where $\Omega$ is the discrete image domain, and $\lambda_{\mathcal{R}}$ is a relative weight.
Since the goal is to learn registration of images with approximately the same anatomy, we train the CNN with pairs of images that are similar to each other -- specifically, within 3 neighbours in the image stack. In order to prevent overfitting, which may be problematic due to the relatively limited number of combinations of pairs, we use random spatial transformations for data augmentation at each iteration in the source and target images, including small random similarity transforms and smooth nonlinear deformations. Once this CNN has been trained, its weights $\bm{\theta}_{\mathcal{R}}$ are frozen during training of the rest of layers in our framework.

\subsection{Image-to-Image translation using a registration loss}

The modality translation is performed by a generator network, $\mathcal{G}$, with a similar architecture to~\cite{park2020contrastive} and trained using a combination of two losses: $\mathcal{L}_{\text{reg}}$ and $\mathcal{L}_{\text{geo}}$. %
The first component $\mathcal{L}_{\text{reg}}$ is the registration loss  between the target and the translated, deformed source.
In Section~\ref{sec:intramodality} above, we used the LNCC metric, which is known to work well in learning-based, intra-modality registration registration of most modalities, and can handle bias field in MRI~\cite{balakrishnan2019voxelmorph}. However, in I2I we  need to explicitly penalise absolute intensity differences, since encouraging local correlation is not enough to optimise the synthesis. For this purpose, we use the  $\ell_1$-norm, which has been widely used in the synthesis literature, and which is more robust than $\ell_2$ against violations of the assumption that the anatomy is perfectly paired in the source and target images. The registration loss is:
\begin{equation}
    \mathcal{L}_{\text{reg}}(S,T; \bm{\theta}_{\mathcal{G}}, \bm{\theta}_{\mathcal{R}}) = \frac{1}{|\Omega|}\sum_{\bm{x}\in\Omega}\| T(\bm{x}) -  S^T(\Phi(\bm{x}; \bm{\theta}_{\mathcal{R}});\bm{\theta}_{\mathcal{G}}) \|_1.
    \label{eq:reg_loss}
\end{equation}

The second component of the loss $\mathcal{L}_{\text{geo}}$ seeks to enforce geometric consistency in the synthesis and is based on noise contrastive estimation (PatchNCE~\cite{park2020contrastive}). The idea behind PatchNCE is to maximise a lower bound on the MI between the pre- and post-synthesis images at the patch level. For this purpose, we define a ``query'' image $q$ (e.g., $S^T$) and a ``reference'' image $r$ (e.g., $S$), from which we extract patch descriptors from the stack of features computed by the encoding branch of the I2I CNN, $\mathcal{G}$.
These descriptors are the output of $L$ layers of interest, including: 
the input image, the downsampling convolutional layers and the first and last ResNet blocks.
Specifically, we extract sets of features $\bm{f}_l$ at the layers of interest $l=1,\ldots,L$ and $N$  random locations $\bm{x}_{l,n}$ per layer, i.e., $\{\bm{f}_{l}(\bm{x}_{l,n})\}_{l=1,\ldots,L; n=1,\ldots,N}$ (in practice, a tissue mask is used when drawing $\bm{x}_{l,n}$ in order not to sample the background).
Each of these $\bm{f}_{l}$ encodes different image features (with different number of channels), from different neighbourhoods (patches), and at different resolution levels.  

Given these descriptors, the contrastive loss builds on the principle that  $\bm{f}^q_{l}(\bm{x}_{l,n})$ (for the query) and $\bm{f}^r_{l}(\bm{x}_{l,n'})$ (for the reference) should be similar for $n=n'$ and dissimilar for $n \neq n'$. Rather than using the descriptors $\bm{f}$ directly, we follow  in~\cite{park2020contrastive} and run them through two-layer perceptrons (which are different for the descriptors in every layer $l$, since they have different resolutions), followed by unit-norm normalisation layers. This yields a new representation $\{\bm{z}_{l,n}\}_{l=1,\ldots,L; n=1,\ldots,N}$, with:
\begin{equation}
\bm{z}_{l,n} = \mathcal{Q}_l(\bm{f}_{l}(\bm{x}_{l,n}); \bm{\theta}_z),
\end{equation}
where $\bm{\theta}_z$ groups the parameters of these representation layers. Given $\bm{z}$, the contrastive PatchNCE loss is given by a softmax function of cosine similarities:
\begin{equation}
    \mathcal{L}_{\text{PatchNCE}}(q,r; \bm{\theta}_z, \tau) = - \frac{1}{N} \sum_{n=1}^N  \sum_{l=1}^{L}\log\left( \frac{\exp(\bm{z}_{l,n}^{q}\cdot \bm{z}_{l,n}^{r}/ \tau)}{ \sum_{n'=1}^{N} \exp(\bm{z}_{l,n}^q \cdot \bm{z}_{l,n'}^{r}/ \tau)}\right),
\end{equation}
where $\tau$ is a temperature parameter and $(\cdot)$ is the dot product. It can be shown that the lower bound on the MI becomes tighter with increasing $N$~\cite{oord2018representation}. 

In practice, we use two PatchNCE losses: one between the source and translated images; and another between the registered and target images:
\begin{align}
\mathcal{L}_{\text{geo}}(S, T; \bm{\theta}_{\mathcal{G}},\bm{\theta}_{\mathcal{R}},\bm{\theta}_z, \tau) = &  
\mathcal{L}_{\text{PatchNCE}}(S^T(\bm{x};\bm{\theta}_{\mathcal{G}}), S(\bm{x}); \theta_z,\tau) \nonumber \\
+ &  
\mathcal{L}_{\text{PatchNCE}}(S^T(\Phi(\bm{x};
\bm{\theta}_{\mathcal{R}});\bm{\theta}_{\mathcal{G}}), T(\bm{x}); \theta_z,\tau). \label{eq:geo_loss}
\end{align}
Combining the registration and geometric consistency losses in Equations~\ref{eq:reg_loss} and~\ref{eq:geo_loss} yields the final loss for our meta-architecture:
\begin{align}
\mathcal{L}(\bm{\theta}_{\mathcal{G}}, \bm{\theta}_z) = & 
\frac{1}{|\Omega|} \sum_{\bm{x}\in\Omega}\| T(\bm{x}) -  S^T(\Phi(\bm{x}; \bm{\theta}_{\mathcal{R}});\bm{\theta}_{\mathcal{G}}) \|_1 \nonumber \\
+ &  
\lambda_{\text{geo}} \mathcal{L}_{\text{PatchNCE}}(S^T(\bm{x};\bm{\theta}_{\mathcal{G}}), S(\bm{x}); \theta_z,\tau) \nonumber \\
+ & 
\lambda_{\text{geo}} \mathcal{L}_{\text{PatchNCE}}(S^T(\Phi(\bm{x};
\bm{\theta}_{\mathcal{R}});\bm{\theta}_{\mathcal{G}}), T(\bm{x}); \theta_z,\tau), 
\end{align}
which we optimise with respect to $\bm{\theta}_{\mathcal{G}}$ and $\bm{\theta}_z$ -- since $\tau$ is a fixed hyperparameter and $\bm{\theta}_{\mathcal{R}}$ is frozen, as explained above.

\section{Experiments and results}

\subsection{Data}
We validate the presented methodology in the context of 3D histology reconstruction via registration to a reference MRI volume. We use two publicly available datasets with histological sections and an \emph{ex vivo} 3D MRI of the same subject. A  3D similarity transform between the stack of histological sections and the MRI volume was used to align images from both domains~\cite{modat2010fast}. The MRI volume was then resampled into the space of histological stack, which yields a set of paired images to register: histological sections and corresponding MRI resampled planes.
The two datasets  are:
\begin{itemize}
    \item \textbf{Allen Human Brain Atlas}~\cite{ding2016comprehensive}: 
    this dataset includes 93 sections with manual delineations of hundreds of brain structures, which we grouped into four coarse tissue classes: cerebral white matter (WM), cerebral grey matter (GM), cerebellar white matter (WMc), and cerebellar grey matter (GMc). An \emph{ex vivo} MRI is available, which was segmented into the same four tissue classes with SPM~\cite{ashburner2005unified}. In addition, J.E.I. manually annotated $13.8 \pm 4.4$ pairs of matching landmarks in the histological sections and corresponding resampled MRI planes, uniformly distributed across all spatial locations.
    
    \item \textbf{BigBrain Initiative}~\cite{amunts2013bigbrain}: we considered one every 20 sections, i.e., one section every 0.4 mm (344 sections in total).
    As in the previous dataset, an \emph{ex vivo} MRI is available and J.E.I. manually annotated $11.6 \pm 1.7$ landmark pairs in the histological sections and corresponding  MRI planes. No segmentations are available for this dataset.
\end{itemize}

\subsection{Experimental setup}
In our experiments, we register each histological section to the corresponding (resampled) MRI  slice. For quantitative evaluation, we report the average root-mean-squared landmark error (both datasets) and the  Dice score on brain tissue classes (only for the Allen dataset).

Our proposed method, \textbf{SbR}, was trained with the following hyper-parameters:  $\lambda_{\text{geo}}=0.02$, $\tau=0.05$ and $\lambda_{\mathcal{R}}=1$, which were set from a subset of the Allen dataset and used elsewhere. We also tested three other configurations of our method: an ablated version  without the structure preserving constraint, i.e., $\lambda_{\text{geo}}=0$ (\textbf{SbR-N}); fine-tuning the result of \textbf{SbR} by unfreezing the registration parameters (\textbf{SbR-R}); and an extension  (\textbf{SbR-G}) that includes an LSGAN loss~\cite{mao2017least} with a PatchGAN discriminator~\cite{isola2017image} to discriminate between synthesised and target images ($S^T$ and $T$). \textbf{SbR-G} enables us to assess the potential benefits of adding a distribution matching loss in  training.

In addition, we compare our method against a number of other methods, to test differences against: standard registration metrics, other synthesis-based approaches without specific geometric constraints, and supervision with labels and Dice scores. Specifically, the competing methods are: (\textit{i}) \textbf{Linear}, the initial affine registration with NiftyReg~\cite{modat2010fast}; (\textit{ii}) \textbf{NMI}, unsupervised training using normalised mutual information (NMI) with 20 bins on the image intensities; (\textit{iii})~\textbf{NMIw}, weakly supervised training using NMI and an additional Dice loss~\cite{milletari2016v} on the segmentations; (\textit{iv}) \textbf{cGAN}, a CycleGAN~\cite{zhu2017unpaired} approach combined with our registration loss; and (\textit{v}) \textbf{RoT},  the state-of-the-art method presented in~\cite{arar2020unsupervised} that consists of alternating the registration and translation steps.
All learning-based methods above (including ours) use the same architecture  for registration, and also the same nonlinear spatial augmentation scheme (sampling $9\times9\times2$ from zero-mean Gaussians and  upsampling to full resolution)

\subsection{Results}
\begin{figure}[!t]
    \centering
    \includegraphics[width=\linewidth]{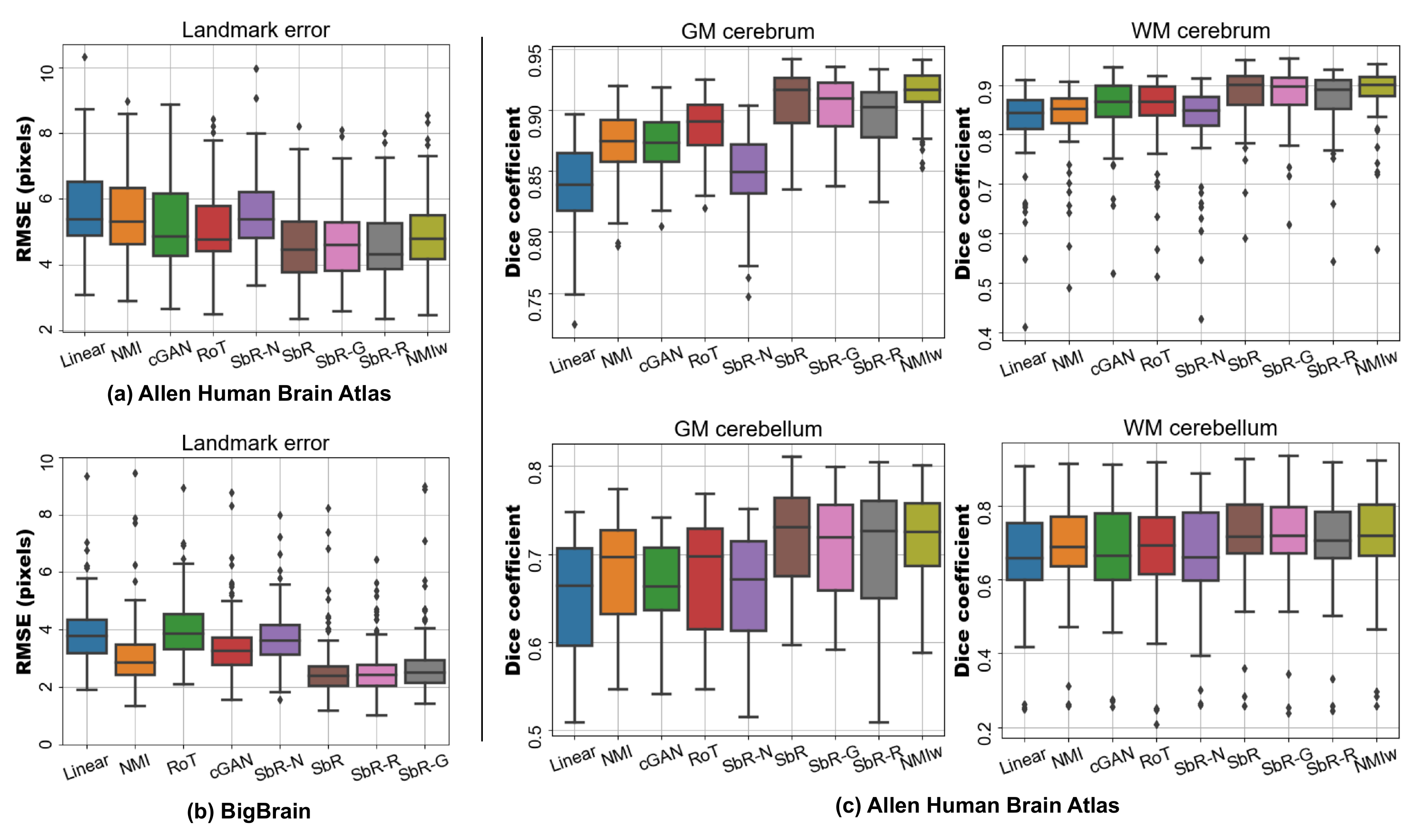}
    \caption{Landmark mean squared error on the Allen human brain atlas dataset (a) and the BigBrain  dataset (b).  Dice score coefficient for the Allen dataset is shown in (c). }
    \label{fig:quantitative_results}
\end{figure}

The quantitative results are summarised in Figure~\ref{fig:quantitative_results}. The landmark errors show that our proposed method (SbR) outperforms all baseline approaches: 
11\%, 9 \% and 7\% error reduction with respect cGAN, RoT and NMIw in the Allen dataset and 23\% and 33 \% with respect cGAN and RoT in the BigBrain dataset; all improvements are statistically significant ($p<0.001$) using a Wilcoxon signed-rank test. Interestingly, SbR is able to align tissue masks as well as NMIw, even though segmentations were not used in the training phase. The naive approach (SbR-N) suffers from synthetic artefacts in the generator, which degrades the results - thus highlighting the importance of including structure preserving constraints in the model. The other two extensions of the model, SbR-G and SbR-R, achieve similar performance to the initial configuration, without yielding any statistically significant additional benefits.

\begin{figure}[!t]
    \centering
    \includegraphics[width=\linewidth]{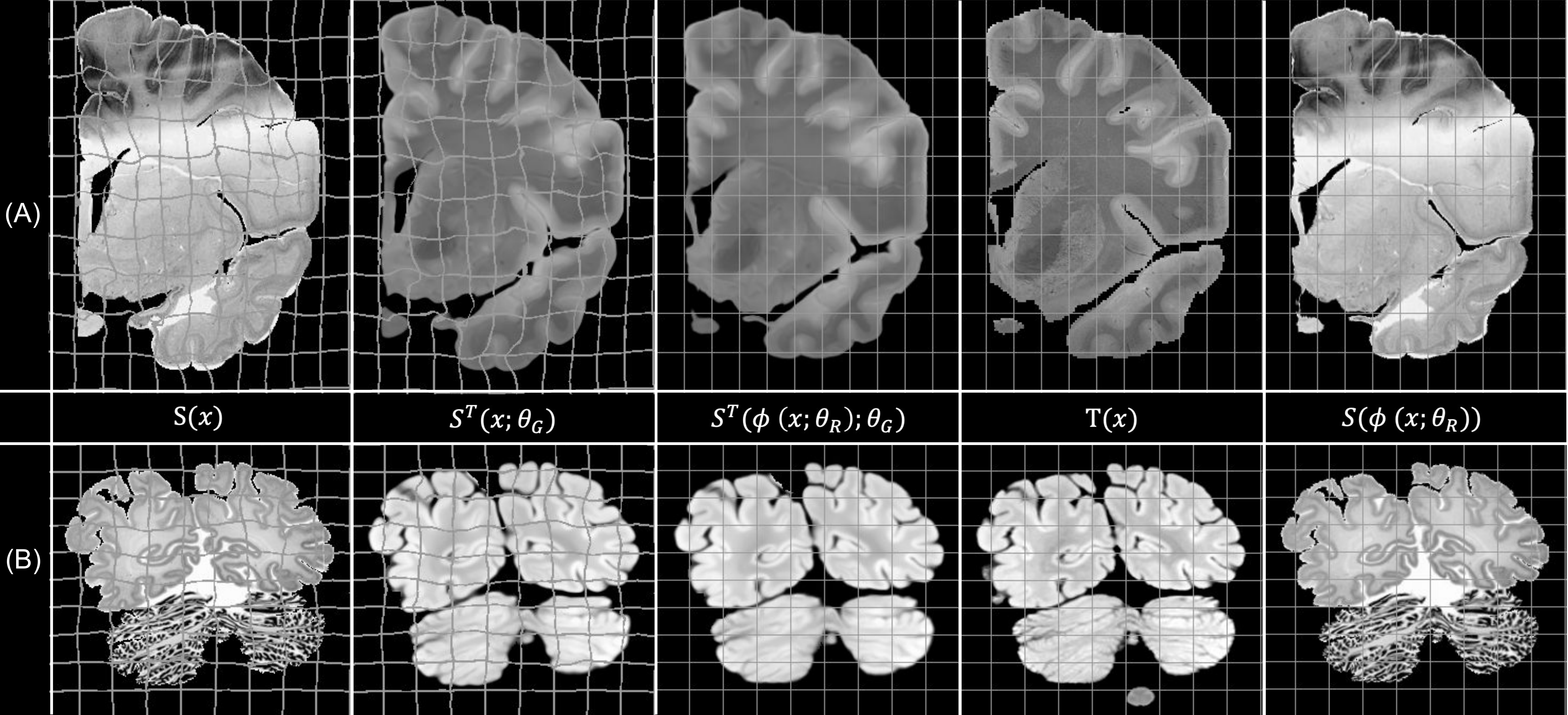}
    \caption{Image examples from (a) the Allen human brain atlas,  and (b) the BigBrain project, with the deformed and rectangular grid overlaid on the source and target spaces, respectively.}
    \label{fig:synthesis_example}
\end{figure}

\begin{figure}[!t]
    \centering
    \includegraphics[width=\linewidth]{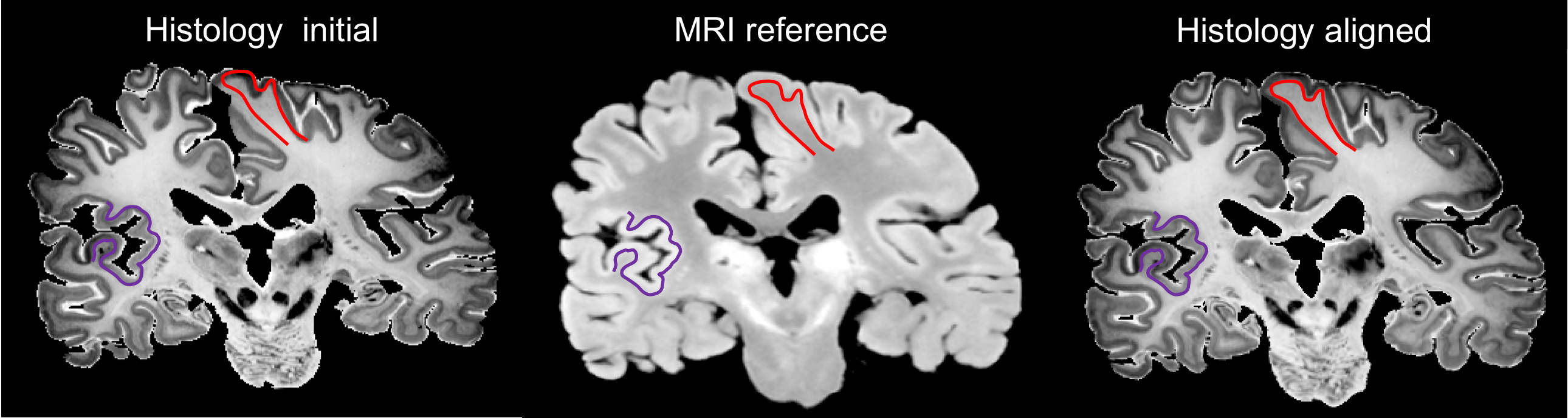}
    \caption{Section 170 from  BigBrain, with cortical boundaries manually traced on the target domain (MRI) and overlaid on the histology, before and after registration.}
    \label{fig:cortex_alignment}
\end{figure}

In Figure~\ref{fig:synthesis_example}, we show an example of the synthesised and registered images using SbR for each dataset. The method displays robustness against common  artefacts, such as: cracks, missing tissue and inhomogeneous staining (in histology), or intensity inhomogeneity (in MRI). Our method is able to accurately register convoluted structures such as the cortex, as seen in Figure~\ref{fig:cortex_alignment}

\section{Discussion and conclusion}
We have presented Synth-by-Reg, a synthesis-by-registration framework for inter-modality registration, which we have validated on a histology-to-MRI registration task. The method uses a single I2I translation network trained with a robust  registration loss (based on the $\ell_1$-norm) and a geometric consistency term (based on contrastive learning). In histology-MRI registration, Synth-by-Reg enables us to avoid using a CycleGAN approach, which often falters in presence of histological artefacts -- since it needs to learn to simulate them and subsequently recover from them.  Future work will focus on adapting our method to the unpaired scenario, as well as to other imaging modalities. We believe that synthesis-by-registration can be a very useful alternative in difficult inter-modality registration problems when weakly paired data are available, e.g.,  MRI and histology. 




\bibliographystyle{splncs04}
\bibliography{refs}

\begin{thebibliography}{10}
\providecommand{\url}[1]{\texttt{#1}}
\providecommand{\urlprefix}{URL }
\providecommand{\doi}[1]{https://doi.org/#1}

\bibitem{amunts2013bigbrain}
Amunts, K., Lepage, C., Borgeat, L., Mohlberg, H., Dickscheid, T., Rousseau,
  M.{\'E}., Bludau, S., Bazin, P.L., Lewis, L.B., et~al.: {BigBrain}: an
  ultrahigh-resolution {3D} human brain model. Science  \textbf{340}(6139),
  1472--1475 (2013)

\bibitem{arar2020unsupervised}
Arar, M., Ginger, Y., Danon, D., Bermano, A.H., Cohen-Or, D.: Unsupervised
  multi-modal image registration via geometry preserving image-to-image
  translation. In: CVPR. pp. 13410--13419. IEEE (2020)

\bibitem{arjovsky2017towards}
Arjovsky, M., Bottou, L.: Towards principled methods for training generative
  adversarial networks. arXiv preprint arXiv:1701.04862  (2017)

\bibitem{ashburner2000voxel}
Ashburner, J., Friston, K.: Voxel-based morphometry-the methods. Neuroimage
  \textbf{11}(6),  805--821 (2000)

\bibitem{ashburner2005unified}
Ashburner, J., Friston, K.: Unified segmentation. Neuroimage  \textbf{26},
  839--851 (2005)

\bibitem{balakrishnan2019voxelmorph}
Balakrishnan, G., Zhao, A., Sabuncu, M.R., Guttag, J., Dalca, A.V.: Voxelmorph:
  a learning framework for deformable medical image registration. IEEE
  transactions on medical imaging  \textbf{38}(8),  1788--1800 (2019)

\bibitem{cao2017deformable}
Cao, X., Yang, J., Zhang, J., Nie, D., Kim, M., Wang, Q., Shen, D.: Deformable
  image registration based on similarity-steered {CNN} regression. In: MICCAI.
  pp. 300--308. Springer (2017)

\bibitem{cciccek20163d}
{\c{C}}i{\c{c}}ek, {\"O}., Abdulkadir, A., Lienkamp, S.S., Brox, T.,
  Ronneberger, O.: {3D U-Net}: learning dense volumetric segmentation from
  sparse annotation. In: MICCAI. pp. 424--432. Springer (2016)

\bibitem{cohen2018distribution}
Cohen, J.P., Luck, M., Honari, S.: Distribution matching losses can hallucinate
  features in medical image translation. In: MICCAI. pp. 529--536. Springer
  (2018)

\bibitem{dalca2018unsupervised}
Dalca, A.V., Balakrishnan, G., Guttag, J., Sabuncu, M.R.: Unsupervised learning
  for fast probabilistic diffeomorphic registration. In: International
  Conference on MICCAI. pp. 729--738. Springer (2018)

\bibitem{ding2016comprehensive}
Ding, S.L., Royall, J.J., Sunkin, S.M., Ng, L., Facer, B.A., Lesnar, P.,
  Guillozet-Bongaarts, A., McMurray, B., et~al.: Comprehensive
  cellular-resolution atlas of the adult human brain. Journal of Comparative
  Neurology  \textbf{524}(16),  3127--3481 (2016)

\bibitem{fan2019adversarial}
Fan, J., Cao, X., Wang, Q., Yap, P.T., Shen, D.: Adversarial learning for
  mono-or multi-modal registration. Medical image analysis  \textbf{58},
  101545 (2019)

\bibitem{fonov2011unbiased}
Fonov, V., Evans, A.C., Botteron, K., Almli, C.R., McKinstry, R.C., Collins,
  D.L.: Unbiased average age-appropriate atlases for pediatric studies.
  NeuroImage  \textbf{54}(1),  313--327 (2011)

\bibitem{heinrich2012mind}
Heinrich, M.P., Jenkinson, M., Bhushan, M., Matin, T., Gleeson, F.V., Brady,
  M., Schnabel, J.A.: {MIND}: Modality independent neighbourhood descriptor for
  multi-modal deformable registration. Medical image analysis  \textbf{16}(7),
  1423--1435 (2012)

\bibitem{hu2012mr}
Hu, Y., Ahmed, H.U., Taylor, Z., Allen, C., Emberton, M., Hawkes, D., Barratt,
  D.: {MR} to ultrasound registration for image-guided prostate interventions.
  Medical image analysis  \textbf{16}(3),  687--703 (2012)

\bibitem{huo2018adversarial}
Huo, Y., Xu, Z., Bao, S., Assad, A., Abramson, R.G., Landman, B.A.: Adversarial
  synthesis learning enables segmentation without target modality ground truth.
  In: ISBI. pp. 1217--1220. IEEE (2018)

\bibitem{iglesias2013synthesizing}
Iglesias, J.E., Konukoglu, E., Zikic, D., Glocker, B., Van~Leemput, K., Fischl,
  B.: Is synthesizing mri contrast useful for inter-modality analysis? In:
  MICCAI. pp. 631--638. Springer (2013)

\bibitem{iglesias2015multi}
Iglesias, J.E., Sabuncu, M.R.: Multi-atlas segmentation of biomedical images: a
  survey. Medical image analysis  \textbf{24}(1),  205--219 (2015)

\bibitem{isola2017image}
Isola, P., Zhu, J.Y., Zhou, T., Efros, A.A.: Image-to-image translation with
  conditional adversarial networks. In: CVPR. pp. 1125--1134. IEEE (2017)

\bibitem{kwon2013portr}
Kwon, D., Niethammer, M., Akbari, H., Bilello, M., Davatzikos, C., Pohl, K.M.:
  {PORTR}: Pre-operative and post-recurrence brain tumor registration. IEEE
  transactions on medical imaging  \textbf{33}(3),  651--667 (2013)

\bibitem{maes2003medical}
Maes, F., Vandermeulen, D., Suetens, P.: Medical image registration using
  mutual information. Proceedings of the IEEE  \textbf{91}(10),  1699--1722
  (2003)

\bibitem{mao2017least}
Mao, X., Li, Q., Xie, H., Lau, R.Y., Wang, Z., Paul~Smolley, S.: Least squares
  generative adversarial networks. In: CVPR. pp. 2794--2802. IEEE (2017)

\bibitem{milletari2016v}
Milletari, F., Navab, N., Ahmadi, S.A.: V-net: Fully convolutional neural
  networks for volumetric medical image segmentation. In: 3DV Conf. pp.
  565--571 (2016)

\bibitem{modat2010fast}
Modat, M., Ridgway, G.R., Taylor, Z.A., Lehmann, M., Barnes, J., Hawkes, D.J.,
  Fox, N.C., Ourselin, S.: Fast free-form deformation using graphics processing
  units. Computer methods and programs in biomedicine  \textbf{98}(3),
  278--284 (2010)

\bibitem{oord2018representation}
Oord, A.v.d., Li, Y., Vinyals, O.: Representation learning with contrastive
  predictive coding. arXiv preprint arXiv:1807.03748  (2018)

\bibitem{park2020contrastive}
Park, T., Efros, A.A., Zhang, R., Zhu, J.Y.: Contrastive learning for unpaired
  image-to-image translation. In: ECCV. pp. 319--345. Springer (2020)

\bibitem{pichat2018survey}
Pichat, J., Iglesias, J.E., Yousry, T., Ourselin, S., Modat, M.: A survey of
  methods for {3D} histology reconstruction. Medical image analysis
  \textbf{46},  73--105 (2018)

\bibitem{qin2019unsupervised}
Qin, C., Shi, B., Liao, R., Mansi, T., Rueckert, D., Kamen, A.: Unsupervised
  deformable registration for multi-modal images via disentangled
  representations. In: IPMI. pp. 249--261. Springer (2019)

\bibitem{reuter2012within}
Reuter, M., Schmansky, N.J., Rosas, H.D., Fischl, B.: Within-subject template
  estimation for unbiased longitudinal image analysis. Neuroimage  \textbf{61},
   1402--18 (2012)

\bibitem{rohlfing2004evaluation}
Rohlfing, T., Brandt, R., Menzel, R., Maurer~Jr, C.R.: Evaluation of atlas
  selection strategies for atlas-based image segmentation with application to
  confocal microscopy images of bee brains. NeuroImage  \textbf{21}(4),
  1428--1442 (2004)

\bibitem{sokooti2017nonrigid}
Sokooti, H., De~Vos, B., Berendsen, F., Lelieveldt, B.P., I{\v{s}}gum, I.,
  Staring, M.: Nonrigid image registration using multi-scale {3D} convolutional
  neural networks. In: MICCAI. pp. 232--239. Springer (2017)

\bibitem{sotiras2013deformable}
Sotiras, A., Davatzikos, C., Paragios, N.: Deformable medical image
  registration: {A} survey. IEEE transactions on medical imaging
  \textbf{32}(7),  1153--1190 (2013)

\bibitem{tanner2018generative}
Tanner, C., Ozdemir, F., Profanter, R., Vishnevsky, V., Konukoglu, E., Goksel,
  O.: Generative adversarial networks for {MR-CT} deformable image
  registration. arXiv preprint arXiv:1807.07349  (2018)

\bibitem{de2017end}
de~Vos, B.D., Berendsen, F.F., Viergever, M.A., Staring, M., I{\v{s}}gum, I.:
  End-to-end unsupervised deformable image registration with a convolutional
  neural network. In: International Workshop DLMIA, pp. 204--212. Springer
  (2017)

\bibitem{wang2021dicyc}
Wang, C., Yang, G., Papanastasiou, G., Tsaftaris, S.A., Newby, D.E., Gray, C.,
  et~al.: {DiCyc}: {GAN}-based deformation invariant cross-domain information
  fusion for medical image synthesis. Information Fusion  \textbf{67},
  147--160 (2021)

\bibitem{wei2019synthesis}
Wei, D., Ahmad, S., Huo, J., Peng, W., Ge, Y., Xue, Z., Yap, P.T., Li, W.,
  Shen, D., Wang, Q.: Synthesis and inpainting-based {MR-CT} registration for
  image-guided thermal ablation of liver tumors. In: MICCAI. pp. 512--520.
  Springer (2019)

\bibitem{xu2020adversarial}
Xu, Z., Luo, J., Yan, J., Pulya, R., Li, X., Wells, W., Jagadeesan, J.:
  Adversarial uni-and multi-modal stream networks for multimodal image
  registration. In: MICCAI. pp. 222--232. Springer (2020)

\bibitem{zhu2017unpaired}
Zhu, J.Y., Park, T., Isola, P., Efros, A.A.: Unpaired image-to-image
  translation using cycle-consistent adversarial networks. In: CVPR. pp.
  2223--2232. IEEE (2017)

\end{thebibliography}

\end{document}